\documentclass{esannV2}
\usepackage[dvips]{graphicx}
\usepackage[latin1]{inputenc}
\usepackage{epstopdf} 
\usepackage{amssymb,amsmath,array}
\usepackage{url}
\graphicspath{ {./} }
\newlength{\bibitemsep}
\newlength{\bibparskip}
\let\oldthebibliography\thebibliography
\renewcommand\thebibliography[1]{%
  \oldthebibliography{#1}%
  \setlength{\parskip}{\bibitemsep}%
  \setlength{\itemsep}{\bibparskip}%
}
\usepackage{placeins}
\usepackage{float}
\usepackage{eso-pic}

\AtBeginDocument{\AddToShipoutPictureFG*{\AtTextUpperLeft{\put(0,\LenToUnit{40pt}){\parbox{\textwidth}{\centering\bfseries
					28th European Symposium on Artificial Neural Networks, Computational Intelligence and Machine Learning (ESANN), Bruges, Belgium, 2020
}}}}}

\begin{document}
\title{Motion Segmentation using \\Frequency Domain Transformer Networks}

\author{Hafez Farazi and Sven Behnke
%
%
\vspace{.3cm}\\
%
University of Bonn, Computer Science Institute VI, Autonomous Intelligent Systems\\
Endenicher Allee 19a, 53115 Bonn, Germany \\
\{farazi, behnke\}@ais.uni-bonn.de
}

\maketitle

\begin{abstract}
Self-supervised prediction is a powerful mechanism to learn representations that capture the underlying structure of the data. Despite recent progress, the self-supervised video prediction task is still challenging. One of the critical factors that make the task hard is motion segmentation, which is segmenting individual objects and the background and estimating their motion separately. In video prediction, the shape, appearance, and transformation of each object should be understood only by predicting the next frame in pixel space. To address this task, we propose a novel end-to-end learnable architecture that predicts the next frame by modeling foreground and background separately while simultaneously estimating and predicting the foreground motion using Frequency Domain Transformer Networks. Experimental evaluations show that this yields interpretable representations and that our approach can outperform some widely used video prediction methods like Video Ladder Network and Predictive Gated Pyramids on synthetic data.
\end{abstract}
\section{Introduction}

Many of the recent models for video prediction use a huge number of parameters, which results in scalability issues and lack of interpretability. Furthermore, these large networks take days to train on even synthetic datasets, which makes exploring new ideas more difficult in comparison with lightweight differentiable models, which only need minutes for training. More importantly, due to the high number of parameters used in heavy models, they tend to overfit the training set and do not easily generalize to novel data. 
One way to address these issues it to prestructure the models based on domain knowledge. Of course, manually engineering every aspect of video prediction is not possible, and one has to find a good balance between nature---inductive bias, which is optimized on an evolutionary time scale---and nurture---learning from own experience. 

In this work, we propose a model for motion segmentation that has zero trainable parameters and is fully interpretable.
It models foreground and background separately. Our model estimates and predicts foreground motion using Frequency Domain Transformer Networks~\cite{Farazi2019}.
We extend this model by adding a few learnable parameters. The improvements made by the added parameters are fully explainable and rational.
The code and dataset of this paper are publicly available.\footnote{ \url{https://github.com/AIS-Bonn/MotionSegmentation}.}
\section{Related Work}
Despite much progress in the field, self-supervised video prediction is still a challenging task. One fundamental issue in video prediction is that the predictor has to segment the scene into individual objects and background and to infer corresponding motions. One attempt to address segmentation of static images is Tagger~\cite{greff2016tagger}. The Tagger network learns to group the representations of different objects and backgrounds iteratively in a self-supervised way. Hsieh et al.~\cite{hsieh2018learning} proposed Spatial Transformer Network~\cite{jaderberg2015spatial} to decompose video frames into individual objects and model their motion separately. Other works do not explicitly model moving segments and rely on unstructured recurrent models to learn these bindings. For example, Cricri et al.~\cite{cricri2016video} added recurrent lateral connections in Ladder Networks to capture the temporal dynamics of video. Recurrent connections and lateral shortcuts relive the deeper layers from modeling spatial detail. The VLN network achieves competitive results to Video Pixel Networks, the state-of-the-art on Moving MNIST dataset, using fewer parameters.

Some works try to learn image relations by separating content and transformation. For instance, PGP~\cite{michalski2014modeling}, which is based on a gated autoencoder model~\cite{memisevic2013learning}, has the assumption that two temporally consecutive frames can be modeled as a linear transformation of each other. In the PGP model, by using a bi-linear model, the hidden layer of mapping units encodes the transformation. These transformation encodings are then used in a hierarchy to predict the next frame. Conv-PGP~\cite{conv-pgp} significantly reduces the number of parameters by utilizing convolutional layers. When predicting video that has location-dependent features, Azizi et al.~\cite{AziziFarazi2018} proposed location-dependent convolutional layers that can model, for example, bouncing on the borders. 

Another related work is Predictive-Corrective networks~\cite{dave2017predictive}, which sequentially make top-down predictions and then correct those predictions with bottom-up observations for the action recognition task. This model adaptively focuses on surprising images where predictions require significant corrections. More recently, Hur et al.~\cite{hur2019iterative} proposed the Iterative Residual Refinement network for jointly predicting optical flow and estimating occlusions.

\vspace{-5px}
\section{Motion Segmentation Network}

\begin{figure}[tb]
\centering
{\includegraphics[width=0.99\textwidth]{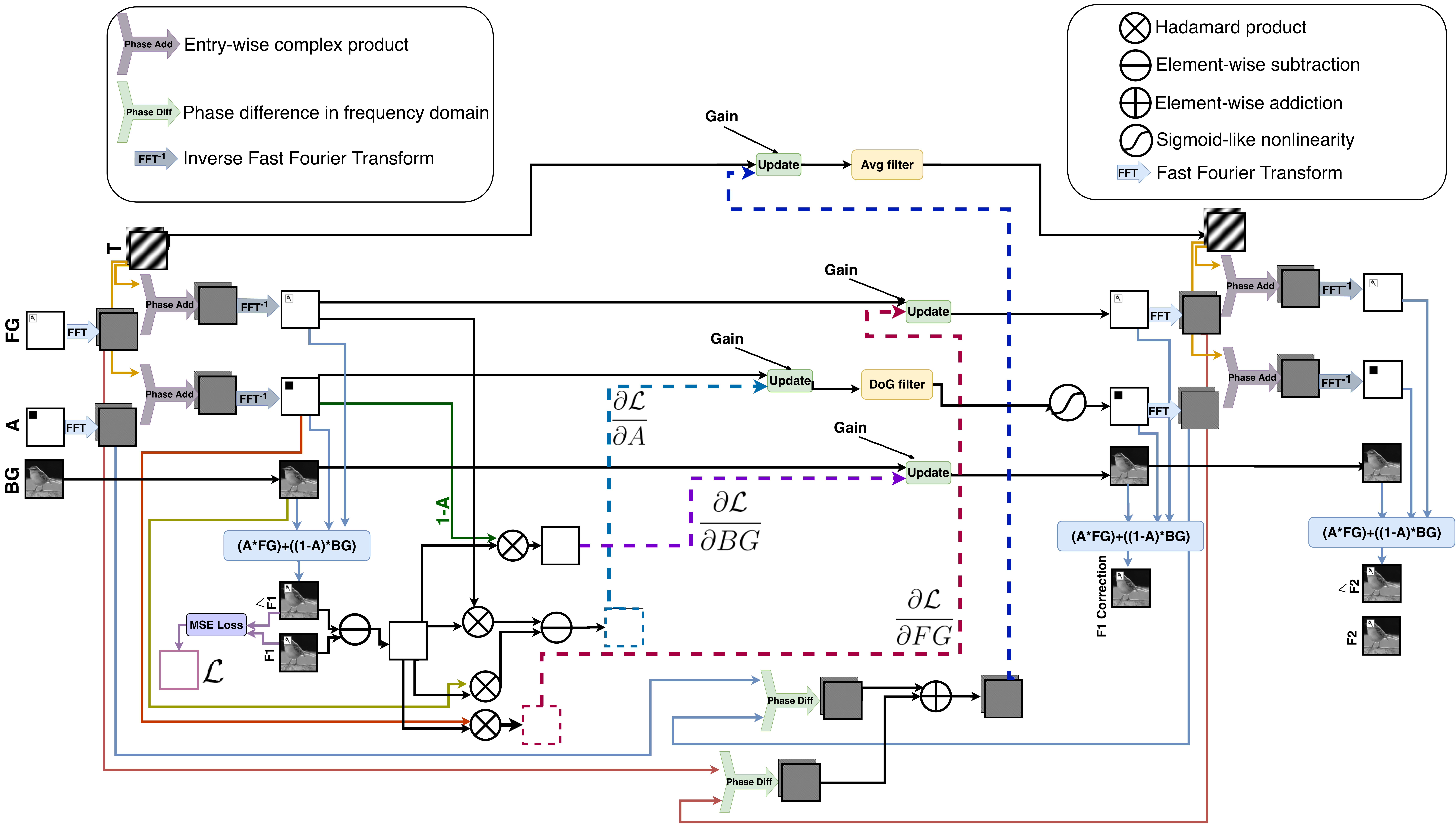}}   
\vspace{-6px}
\caption{Motion segmentation model. Foreground ({\sf FG}) and background ({\sf BG}) are modeled separately and combined using an alpha mask ({\sf A}) to the predicted frame $\hat{{\sf \textrm{F1}}}$, which is compared to the input frame {\sf F1}. The prediction error is used to update {\sf FG}, {\sf BG}, and {\sf A}. 
For {\sf FG} and {\sf A}, motion is estimated by computing phase differences in Fourier space ({\sf T}). This motion estimate is added to the phases of {\sf FG} and {\sf A} to move them accordingly. After a few steps of this prediction-correction cycle, the model does not need the input frames anymore and can continue predicting using only the estimated state ({\sf FG}, {\sf BG}, {\sf A}, {\sf T}).}
\label{Teaser}
\vspace{-12px}
\end{figure}

\subsection{Prediction-correction State Estimation}
Similar to \cite{dave2017predictive}, we were inspired by classic linear dynamical systems theory and Kalman filters. In a Kalman filter, $x_t$ is a noisy linear function of the previous time step state $x_{t-1}$. The observation $z_t$ is modeled as a noisy linear function of the state $x_t$: \vspace*{-1ex}
\begin{equation}
\begin{array}{l}{\mathbf{x}_{t}=\mathbf{F} \mathbf{x}_{t-1}+\text {Noise}} \\ {\mathbf{z}_{t}=\mathbf{H} \mathbf{x}_{t}+\text {Noise}}\end{array},
\vspace{-5px}
\end{equation}
where $F$ is the state-transition matrix, and $H$ is the measurement matrix. 
Under these assumptions, the posterior estimate of the state $x_t$ is calculated by: 
\begin{equation}
\hat{\mathbf{x}}_{t}=\underbrace{\hat{\mathbf{x}}_{t | t-1}}_{\text {prediction }}+\,\, \underbrace{\mathbf{K}(\mathbf{z}_{t}-\hat{\mathbf{z}}_{t | t-1}}_{\text {correction }}),
\end{equation}
where $\hat{x}_{t | t-1}$ and $\hat{z}_{t | t-1}$ are the predictions of $x_t$ and $z_t$, respectively, given observations $z_1,\ldots , z_{t-1}$. $K$ is the Kalman gain matrix, which controls how much we rely on the current prediction $\hat{x}_{t | t-1}$ versus the observation $z_t$.

\subsection{Frequency Domain Motion Segmentation}

Fig.~\ref{Teaser} illustrates our model for self-supervised motion segmentation.
We model foreground (${\sf FG_{\,t}}$) and background (${\sf BG_{\,t}}$) separately as images having the same size as the observed frames (${\sf F_t}$).
Both are combined by modeling occlusion of the background by the foreground using the alpha mask ${\sf \hat{A}_{\,t}}$: 
\begin{equation}{\sf \hat{F}_t} = {\sf \hat{A}_{\,t}} \cdot {\sf \hat{FG}_{\,t}} + (1-{\sf \hat{A}_{\,t}}) \cdot {\sf \hat{BG}_{\,t}}.
\end{equation}
In addition to these three images, the state also consists of the estimated common movement speed ${\sf T_t}$ of  foreground and alpha mask. ${\sf T_t}$ is represented as phase differences (unit length complex numbers) between consecutive frames in the Fourier domain. 
It has the same size as the images. As in the Frequency Domain Transformer Networks~\cite{Farazi2019}, the next foreground frame (${\sf \hat{FG}_{\,t}, \hat{A}_{\,t}}$) can easily be predicted by phase-adding $\sf T_t$ to the Fourier representations $FFT(.)$ of ($\sf FG_{\,t-1}, \sf A_{\,t-1}$)
which is realized by element-wise multiplication of these complex matrices. After going back to the spatial domain by the inverse Fourier transformation $FFT^{-1}(.)$, the foreground and alpha mask are moved according to the estimated movement speed.

We calculate the difference between the predicted frame $\sf \hat{F}_t$ and observed frame $\sf F_t$ and update each part of the state to minimize the mean squared loss $\mathcal{L}(\sf \hat{F}_t, F_t)$.
As the predicted frame is computed by a simple differentiable function graph, we can easily perform gradient descent by a function graph for the backward pass that has the same structure. Instead of using automatic differentiation packages for updating each state, we hard-wired gradient computation in our computational graph. This results in a computation graph that realizes a Kalman filter-like prediction-correction cycle in its forward pass.
For updating the state $\sf T_t$, which is in Fourier space, we calculate the phase differences between $FFT(\sf A_{\,t})$ and $FFT(\sf A_{\,t-1})$ as well as $FFT(\sf FG_{\,t})$ and $FFT(\sf FG_{\,t-1})$. For computing $\sf T_t$, we take the weighted average between $\sf T_{t-1}$ and calculated phase differences $\sf \tilde{T}_t$. 

We also include two filtering mechanism for {\sf A} and {\sf T} after each update. With the assumption of blob-like response in {\sf A}, we apply a Difference of Gaussian filter. We also filter the phase difference by an averaging filter, with the assumption that the phase difference between adjacent rows and adjacent columns are near-constant. The effect of removing phase filtering is illustrated in Fig.~\ref{SampleRes}(c).

The proposed model hard-wires our assumptions that the foreground moves in front of a stationary background and that it occludes the background according to the alpha mask. Furthermore, we hard-wire motion estimation and prediction by Frequency Domain Transformer Networks~\cite{Farazi2019}. 

\vspace{-5px}
\subsection{Model Extension by Learnable Layers} 
\vspace{-3px}
So far, our model has zero learnable parameters.
Hence, it hard-wires our assumptions, but cannot learn to exploit the statistical properties of data. 
Since our prediction-correction computation graph is differentiable, we can backpropagate a loss through the network that is unfolded in time. 
Hence, any parameter can be updated by gradient descent, and we can easily add parameters at suitable computation steps.

For initializing the spatial states {\sf FG}, {\sf BG}, and {\sf A} we use three different convolutional networks. Each has four convolutional layers, with DenseNet-like connections, followed by ReLU activations. We also initialize {\sf T} using the first two steps of {\sf A} and {\sf FG}. At each time step, each state is the weighted average between the updated state and the output of the convolutional network. We use a decaying gain for this weighted average so that in the initial step, we only use the convolutional network output, and later we rely more and more on the updated states. Note that the convolutional network also fills-in occluded parts of the background {\sf BG}.
\vspace{-5px}
\section{Experimental Results}
\vspace{-3px}
\subsection{Dataset and Training}

We use a variant of the \textit{Moving MNIST} data set to evaluate our proposed architecture.
It contains twenty frames with one MNIST image, moving inside a 128$\times$128 frame. Foreground moving objects were chosen randomly from training and test set and placed at a random position with a random velocity and random background image chosen from the STL-10 dataset. Note that the objects are moved with subpixel velocity.

The models and the update gains are trained end-to-end using backpropagation through time. We used Adam optimizer and MSE prediction loss as well as the cyclic learning rate.
\vspace{-7px}
\subsection{Evaluation}
\vspace{-5px}
\begin{figure}[tb]
	\vspace{0px}
	\centering
	{\includegraphics[clip,trim=5 10 5 0,width=0.96\textwidth]{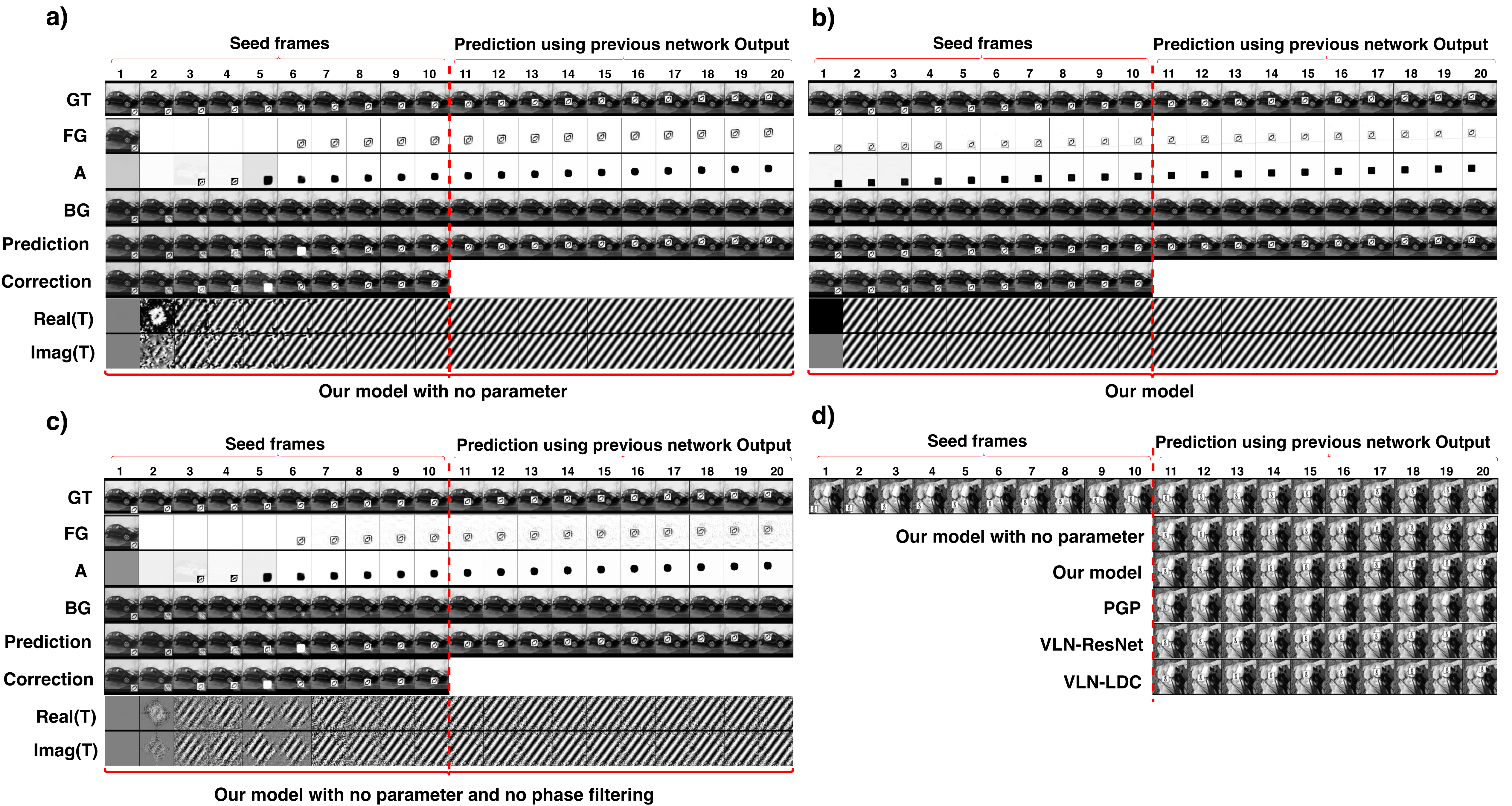}}   
	\caption{ a,b) Internal states' development for one sample Moving MNIST sequence (randomly selected). Note that albeit with varying success levels, both models can segment foreground and background and estimate foreground motion. c) Effect of removing phase filtering. d) Predictions for a randomly selected sample with different models.}
	\label{SampleRes}
	\vspace{-15px}
\end{figure}
We evaluate our architecture against Conv-PGP and two different VLN models. In these experiments, we predicted ten frames from ten seed inputs. Sample results of our models, as well as used baselines, are presented in Fig.~\ref{SampleRes}. 
Fig.~\ref{SampleRes} also shows the development of the internal states of our two model variants for one moving MNIST sample in detail.
The representations are easily interpretable. They segment foreground and background and estimate the foreground speed.

Table~\ref{Table_results} reports the prediction losses, structural similarity, and the number of parameters for the evaluated models. It can be observed that our proposed model outperforms our baselines. 

\begin{table}[tb]
	\vspace{-7px}
	\renewcommand{\arraystretch}{1}
	\centering
	\caption{Prediction losses for Moving MNIST.}
	\vspace*{2px}
	\footnotesize
	\begin{tabular} 
		{l | c | c | c| c| c}
		\footnotesize{Model\hspace{0px}}& \footnotesize{\hspace{0px}L1\hspace{0px}} &  \footnotesize{\hspace{0px}MSE\hspace{0px}}  & \footnotesize{\hspace{0px}SSIM\hspace{0px}} &\footnotesize{\hspace{0px}\# of params\hspace{0px} }\\
		\hline \hline
		\footnotesize{Conv-PGP~\cite{conv-pgp}} &  0.0323   &0.0074&0.9025& 32K\\ 
		\footnotesize{Our model}  & \textbf{0.0024}  & \textbf{0.0002}&\textbf{0.9896}&2K\\
		\footnotesize{Our model without param} & 0.0059  & 0.0010&0.9737&\textbf{0}\\
		\footnotesize{VLN-ResNet~\cite{cricri2016video}} \hspace{3px}&0.0166&0.0009& 0.9540&1.3M\\
		\footnotesize{VLN-LDC~\cite{AziziFarazi2018}} \hspace{3px}&0.0126 &0.0006&0.9686&1.4M\\
		
	\end{tabular}
	\vspace{-3px}\\
	\label{Table_results}
	\vspace{-12px}
\end{table}
\vspace{-2px}
\section{Conclusion}
\vspace{-6px}
We proposed an end-to-end learnable neural network for motion segmentation that models foreground and background separately and predicts foreground motion by Frequency Domain Transformer Networks. 
The network estimates interpretable internal states using a hard-wired prediction-correction scheme.
The basic method is highly computationally efficient and has zero parameters.
We added a few trainable layers to optimize prediction for the specific dataset at hand. 
Experiments indicate that our method can solve the motion segmentation task in synthetic dataset. With far fewer parameters, our proposed architecture significantly outperforms the results of both VLN and Conv-PGP models on the tested dataset. The model with learnable parameters performs better than the model without parameters.

\vspace{-6px}
{\footnotesize \paragraph{Acknowledgment}
\label{acknowledgment}
This work was funded by grant BE 2556/16-1 (Research Unit FOR 2535 Anticipating Human Behavior) of the German Research Foundation (DFG).
}
\vspace{-7px}
\bibliographystyle{unsrt}
{\small \bibliography{document}}

\end{document}